\documentclass[journal,10pt]{IEEEtran}
 
\usepackage{cuted}
\usepackage{comment}
\usepackage{gensymb}
\usepackage{graphicx}
\usepackage{caption}
\usepackage [noadjust]{cite} 
\usepackage{bm}  
\usepackage{amsmath} 
\usepackage{amsfonts} 
\usepackage{amssymb} 
\usepackage{graphicx} 
\usepackage{hyperref}

\usepackage{amsmath}
\usepackage{amssymb}
\usepackage{enumerate}
\usepackage{stfloats}
\usepackage{cases}
\usepackage{epstopdf}
\usepackage{soul,color} 
\usepackage{hyperref}
\usepackage[percent]{overpic}
\usepackage{tabularx}
\usepackage[table]{xcolor}
\usepackage{multirow}
\usepackage{booktabs}  
\usepackage{tabu} 

\usepackage{algorithm}
\usepackage{algpseudocode}%

\newcolumntype{L}[1]{>{\hsize=#1\hsize\raggedright\arraybackslash}X}%
\newcolumntype{R}[1]{>{\hsize=#1\hsize\raggedleft\arraybackslash}X}%
\newcolumntype{C}[1]{>{\hsize=#1\hsize\centering\arraybackslash}X}%

\renewcommand{\mathbf}[1]{\boldsymbol{#1}}

\newcommand{\Ns}{N_{\mathrm{s}}}
\newcommand{\Nb}{N_{\mathrm{b}}}
\newcommand{\Dv}{\mathcal{D}_\mathrm{v}}
\newcommand{\Dvs}{\tilde{\mathcal{D}}_\mathrm{v}}
\newcommand{\Pvs}{\tilde{\boldsymbol{P}}_\mathrm{v}}
\newcommand{\Pv}{\boldsymbol{P}_\mathrm{v}}

\newcommand{\Denv}{\mathcal{D}_\mathrm{e}}
\newcommand{\Denvs}{\tilde{\mathcal{D}}_\mathrm{e}}
\newcommand{\Penv}{\boldsymbol{P}_\mathrm{e}}
\newcommand{\Penvs}{\tilde{\boldsymbol{P}}_\mathrm{e}}
\newcommand{\Dsen}{\mathcal{D}_\mathrm{s}}
\newcommand{\Dsens}{\tilde{\mathcal{D}}_\mathrm{s}}
\newcommand{\Psens}{\tilde{\boldsymbol{P}}_\mathrm{s}}
\newcommand{\Psen}{\boldsymbol{P}_\mathrm{s}}
\newcommand{\Ls}{L_\mathrm{s}}
\newcommand{\Vs}{V_\mathrm{s}}
\newcommand{\hs}{h_\mathrm{s}}

\newtheorem {remark}{Remark}

\begin{document}
	\raggedbottom
	\allowdisplaybreaks


   \title{Vessel Length Estimation from Magnetic Wake Signature: A Physics-Informed Residual Neural Network Approach 

}
    \author{Mohammad Amir Fallah, Mehdi Monemi, \textit{Member}, IEEE, Matti Latva-aho, \textit{Fellow}, IEEE 
 \thanks{
  Mohammad Amir Fallah is with Department of Engineering, Payame Noor University (PNU), Tehran, Iran (email: mfallah@pnu.ac.ir)  
  \\
  Mehdi Monemi is with Centre
for Wireless Communications (CWC), University of Oulu, 90570 Oulu, Finland.
 (emails: mehdi.monemi@oulu.fi).
\\
M. Latva-aho is with Centre for Wireless Communications (CWC), University of Oulu, 90570 Oulu, Finland (email: matti.latva-aho@oulu.fi).
\\
}}

	\maketitle
\begin{abstract}



Marine remote sensing enhances maritime surveillance, environmental monitoring, and naval operations. Vessel length estimation, a key component of this technology, supports effective maritime surveillance by empowering features such as vessel classification.
Departing from traditional methods relying on two-dimensional hydrodynamic wakes or computationally intensive satellite imagery, this paper introduces an innovative approach for vessel length estimation that leverages the subtle magnetic wake signatures of vessels, captured through a low-complexity one-dimensional profile from a single airborne magnetic sensor scan. The proposed method centers around our characterized nonlinear integral equations that connect the magnetic wake to the vessel length within a realistic finite-depth marine environment. 
To solve the derived equations,  we initially leverage a deep residual neural network (DRNN). The proposed DRNN-based solution framework is shown to be unable to exactly learn the intricate relationships between parameters when constrained by a limited training-dataset. To overcome this issue, we introduce an innovative approach leveraging a physics-informed residual neural network (PIRNN). This model integrates physical formulations directly into the loss function, leading to improved performance in terms of both accuracy and convergence speed.
Considering a sensor scan angle of less than $15^\circ$, which maintains a reasonable margin below Kelvin's limit angle of $19.5^\circ$, we explore the impact of various parameters on the accuracy of the vessel 
 length estimation, including sensor scan angle, vessel speed, and sea depth. Numerical simulations demonstrate the superiority of the proposed PIRNN method, achieving mean length estimation errors consistently below 5\% for vessels longer than 100m. For shorter vessels, the errors generally remain under 10\%.

	\end{abstract}
	
\thispagestyle{empty}

\section{Introduction}

 Maritime surveillance and vessel traffic control have consistently been of critical importance, propelling the advancement of state-of-the-art remote sensing systems tailored for marine environments. The emergence of technologically advanced marine vessels has highlighted the need to modernize existing detection technologies, including those based on Synthetic Aperture Radar (SAR), lidar, and sonar methodologies. The principal objective of these technologies remains vessel identification through the detection of anomalies, employing two primary approaches: active and passive detection. 
 Active detection methods involve transmitting a signal and analyzing its reflection from the target, while passive methods rely on distinguishing the vessel's signal from the background environmental noise. 
 Active detection methods, including Lidar and SAR, while capable of providing precise target localization, are vulnerable to countermeasures such as jamming and deception, thereby compromising their operational reliability \cite{10664452}. Moreover, these active airborne techniques are typically limited to the detection of surface vessels and are ineffective for detecting submerged targets. Consequently, contemporary research increasingly focuses on the development of passive methods for detecting distant vessels. 
Sonar-based systems have been a popular choice due to their low acoustic path loss in seawater. However, challenges such as interference from unwanted acoustic signals, intentional deception techniques, and advances in stealth vessel technology have limited their effectiveness. 

To address these limitations, researchers have increasingly turned to magnetic anomaly detection (MAD) as a promising alternative \cite{10641654,liu2023theories}. MAD leverages the magnetic field disturbances generated by moving vessels to identify their presence. While traditional MAD methods focus on detecting vessels with metallic hulls, recent advancements in degaussing technologies have necessitated the exploration of new approaches. 
One such approach involves detecting the hydrodynamic anomalies induced by moving vessels \cite{fallah2013electromagnetic,FALLAH201460}. This method focuses on identifying hydrodynamic anomalies known as hydrodynamic wakes \cite{Newman1977}. The hydrodynamic wake has two regimes of near and far field; the later called as Kelvin waves is involved in the detection of the remote vessel \cite{fallah2021optimal,HUANG2024103933,HUANG2024117779}. The movement of water as a weak conductor in the natural magnetic field of the earth induces magnetic anomaly whose behavior is very similar to the original hydrodynamic Kelvin waves and thus can lead to the detection of the remote vessel if precisely measured and processed with a magnetometer \cite{Zou2000,HUI2023}. A key characteristic of magnetic anomaly induced by the moving vessel is that it lasts for a long time and can be sensed at large distances which makes it a good tool as a passive remote detection method in sea \cite{monemi2023novel,fallah2021optimal,HUI2023,CHEN2023103750}. On the other hand, the aforementioned magnetic anomaly does not depend on the vessel body material, and provides the feasibility of detecting remote vessels equipped with degaussing tools or having non-metallic hulls.

From a remote sensing perspective, beyond mere vessel detection, the acquisition of static and dynamic information, such as speed, heading, and hull dimensions, is also of considerable interest. The estimation of vessel speed and heading using magnetic wake analysis has been addressed in prior research \cite{monemi2023novel,fallah2021optimal,FALLAH201460,fallah2013electromagnetic}. However, the estimation of vessel length from hydrodynamic wakes has been a subject of ongoing investigation, with various techniques proposed for extracting this information from remote sensing data. 
Early studies, such as that of \cite{wu1991remote}, focused on analyzing the spectral characteristics of 2D hydrodynamic vessel wakes. By examining parameters like peak frequency, researchers were able to develop theoretical models for vessel length estimation. While these approaches show potential, their accuracy is sensitive to environmental factors, including assumptions of infinite water depth, wind conditions, and sea state, all of which can substantially influence their precision.  
More recent research has leveraged advanced image processing techniques to directly estimate vessel dimensions from satellite imagery. Authors in \cite{stasolla2016exploitation} introduced a method for accurately outlining vessels in Sentinel-1 SAR images. This method, which employed image processing techniques to refine vessel shapes, allowed for precise measurements of vessel length and width. While this approach demonstrated high accuracy, particularly in estimating vessel length, its performance was hindered by limitations in image resolution and the presence of significant sea clutter.
The application of deep learning has significantly advanced the field of vessel size estimation. A deep learning model, known as SSENet , has been specifically designed to directly estimate the size of vessels from Sentinel-1 Synthetic Aperture Radar images \cite{ren2023extracting}. 
While demonstrating impressive performance, the model's dependency on large-scale accurately labeled datasets as well as the high computational complexity of the solution scheme limit its practical application.
Beyond size estimation, SAR imagery has also been explored to extract vessel hull characteristics from their wake patterns \cite{griffin1996ship}. The use of features like the width, length, and intensity of the wake to classify vessel types and estimate their size and speed is investigated.
While SAR imagery offers unique advantages, optical imagery remains a valuable resource for vessel size estimation. A method has been presented to estimate vessel size from satellite optical images by fitting an ellipse to the vessel's outline \cite{park2020estimation}. 

In this work, we focus predominantly on estimating the length of a vessel through the analysis of its magnetic wake. 
Unlike traditional methods that rely on analyzing the  \textbf{2D hydrodynamic wake}, this research exclusively utilizes a one-dimensional data acquisition of the magnetic wake (\textbf{1D magnetic wake})  to determine the vessel length.
The necessary hydromagnetic equations are derived, and the relationship between the magnetic wake and the static and dynamic parameters of the vessel is presented. These parameters include its speed, direction of movement, and physical dimensions such as length and width. This relationship is then expressed mathematically in the form of nonlinear integral equations. 
A deep residual neural network (DRNN) is employed to solve these nonlinear integral equations. 
However, the performance of the DRNN model was found to be \textbf{inferior}. This limitation arises from the intricate nonlinear relationships between the model parameters and the inherent data-driven nature of the DRNN. Consequently, its performance is heavily reliant on the availability of extensive and accurate training data, which can be challenging to obtain due to limitations in both data quantity and quality. To address these limitations, a physics-informed residual neural network (PIRNN) has been employed, Integrating the governing physical laws directly into the learning process. 
This approach ensures that the generated solutions are physically \textbf{realistic} and align with established principles, leading to significantly improved outcomes. The performance of the proposed PIRNN model was rigorously evaluated through simulations. The results demonstrate that this approach overcomes the limitations of previous methods, which often rely on simplified vessel hull representations and assume noise-free conditions\cite{wu1991remote}. This provides a more flexible solution applicable to vessels with complex geometries, realistic noise levels, and finite water depths, where the underlying physics exhibit greater complexity compared to the idealized scenario of infinite water depth. 

\begin{figure}
	\centering
		\includegraphics [width=250pt]{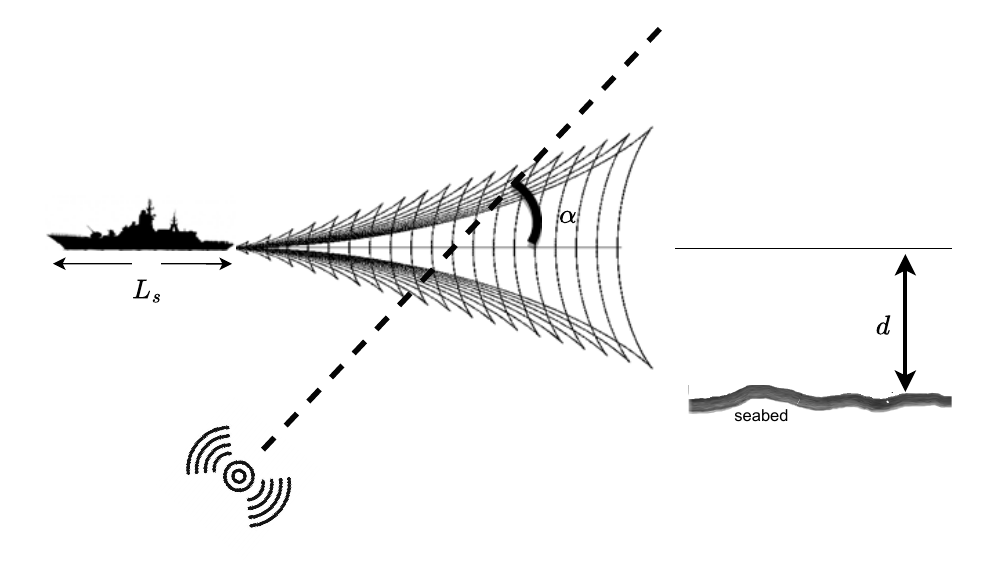} \\
		\caption{Airborne magnetic wake measurement. }
		\label{fig:airborne sensor}
\end{figure}

{\color{black}\subsection{Motivation and Contributions}
\begin{itemize}

\item \textbf{Leveraging magnetic wake in vessel length estimation:}
We introduce a novel approach for estimating vessel length by analyzing its magnetic wake. Conventional methods for determining vessel dimensions often rely on optical or radar imagery, which can be limited by factors such as cloudy weather conditions (restricting satellite-based observations), clutter, and image resolution. The magnetic wake, however, offers a unique and potentially more robust source of information for vessel characterization.
 
\item  \textbf{High speed, low complexity 1D data acquisition:}
Previous studies on vessel length estimation have utilized two-dimensional (2D) data derived from the hydrodynamic vessel wake. 
This study utilizes one-dimensional (1D) data acquired by sampling the two-dimensional magnetic wake field along a \textbf{linear trajectory} traversed by an airborne magnetic sensor. This approach not only simplifies the process of data acquisition but also significantly reduces the sampling time and computational complexity.  

\item  \textbf{Linking vessel magnetic Wake to its length:}
A nonlinear relationship has been derived that intricately connects a vessel's magnetic wake to its physical parameters, including length. This analysis considers the complexities of a finite-depth marine environment, where the underlying physical interactions are more intricate than the simplified assumptions of infinite water depth. 

\item  \textbf{Physics-informed deep learning:}
To estimate vessel parameters (including length) from its magnetic wake, we employed a deep residual neural network, incorporating a loss function that directly integrates with the governing aforementioned physical laws of the problem. This physics-informed deep residual neural network renders the proposed solution practical and feasible for real-world applications. Rigorous simulations demonstrate the efficacy of this approach, with the proposed PIRNN model achieving a vessel length estimation error margin \textbf{up to 5\%.}
\end{itemize}
}
 \subsection{Notations:}Vectors and scalars are denoted by \textbf{bold} and regular letters respectively, $i$ is the unit imaginary number, and $\mathbf{a}_x$, $\mathbf{a}_y$, and $\mathbf{a}_z$ are the unit vectors in the Cartesian coordinate system in the direction of the $x$, $y$, and $z$-axis respectively, and $``\cdot"$ stands for dot-product operation. The notations $\times$ and $\mathbf{\times}$ stand for regular and cross-product operations respectively.
 
\setcounter{equation}{5}
\begin{figure*}[!b]
	\hrulefill
\begin{align}
\label{eq:Maxwell’s equation}
    \mathbf{\nabla} \mathbf{\times} \mathbf{B} = 
    \begin{cases}
        \mu_{\mathrm{a}} \varepsilon_{\mathrm{a}} \frac{\partial \mathbf{E}}{\partial t}, \quad z > 0 \\
        \mu_{\mathrm{w}} \varepsilon_{\mathrm{w}} \frac{\partial \mathbf{E}}{\partial t} + \mu_{\mathrm{w}} \sigma_{\mathrm{w}} (\mathbf{E} + \mathbf{U} \mathbf{\times} \mathbf{B}_T) + \mu_{\mathrm{w}} \rho_{\mathrm{e}} \mathbf{U} 
        + \mu_{\mathrm{w}} (\varepsilon_{\mathrm{w}} - \varepsilon_0) \mathbf{\nabla} \mathbf{\times} (\mathbf{E} + \mathbf{U} \mathbf{\times} \mathbf{B}_T) \mathbf{\times} \mathbf{U}, \quad -d < z < 0 \\
        \mu_{\mathrm{b}} \varepsilon_{\mathrm{b}} \frac{\partial \mathbf{E}}{\partial t} + \mu_{\mathrm{b}} \sigma_{\mathrm{b}} \mathbf{E}, \quad z < -d \\
    \end{cases}
\end{align}
\end{figure*}
\setcounter{equation}{0}

\section{System Model and Problem Formulation}
\label{sec:system_model_and_problem_formulation}
In what follows first we present the system model, and then we express the formulation of the relationship between the hydrodynamic and magnetic expressions necessary for the development of our vessel length estimation scheme. 
Consider a vessel traveling at a uniform speed $V_s$ in the opposite direction to the $x$-axis within a finite-depth seawater, as illustrated in Fig. \ref{fig:airborne sensor}. The sea surface is located on the $xy$-plane, the $z$-axis is perpendicular to the sea surface, and the direction of $y$-axis is found according to the right-hand law. The regions $0<z$, $-d<z<0$, and $z<-d$  correspond to the space above the sea surface, the seawater, and the space under the seabed, respectively. We assume that the geomagnetic field of the earth is constant everywhere.

\subsection{Hydrodynamic Preliminaries}

The movement of an object in a fluid results in the formation of both near and far-field hydrodynamic anomalies. Since we are interested in detecting distant vessels, we only deal with the far-field wakes called Kelvin waves \cite{Newman1977,madurasinghe1994induced}. Let $\phi(x, y, z, t, \theta)$ be the velocity potential function of the fluid (i.e., seawater) corresponding to a wavefront of angle $\theta$ with respect to the $x$-axis. $\phi$ is expressed as \cite{Newman1977}:
\begin{multline}
\label{eq:velocity potential function}
    \phi(x, y, z, t, \theta) =
    \\
    \frac{\tilde{A}({\theta}) g}{\omega_0}\times\frac{\cosh(k_0 (z + d))}{\cosh(k_0 d)}
    \times e^{-i (\omega_0 t + k_0 x \cos(\theta) + k_0 y \sin(\theta))},
\end{multline}
where $g$ is the gravity force, $d$ is the sea depth, $k_0$ is the wavenumber which is obtained from equation $k_0 \tanh(k_0 d) = {\omega_0^2}/{g}$, in which $\omega_0 = k_0 V_s \cos(\theta)$ is the dispersion relation, and finally $\tilde{A}({\theta})$ is obtained from the following equation \cite{FALLAH201460}:
\begin{align}
\tilde{A}({\theta})  = \underbrace{\frac{2 \omega_0 g}{\cos^3(\theta) \pi V_s^3} \frac{e^{k_0 d} - e^{-k_0 d}}{e^{2k_0 d} - e^{-2k_0 d} - 4 k_0 d}}_{A(\theta)} \kappa({\theta}),
\end{align}
where $\kappa({\theta})$ is called the Kochin function defined as follows:
\begin{multline}
\kappa({\theta}) = \iint_S  I_s(x', y', z') e^{-ik_0 (x' \cos(\theta) + y' \sin(\theta))}\\
 \times\cosh(k_0 (z' + d)) \, dx' \, dy'
\end{multline}

and 
\begin{equation}
I_s(x', y', z') = \frac{V_s}{2 \pi} \frac{\partial S}{\partial x}
\end{equation}
represents the intensity of the source located at $(x', y', z')$ and distributed over the wetted part of the submerged portion of the vessel hull. Here, we consider $y = S(x, z)$ as the equation representing the vessel hull geometry.
Once the velocity potential function is obtained from \eqref{eq:velocity potential function}, the fluid velocity vector resulting from the movement of the vessel is calculated as follows:
\begin{equation}
\label{eq:fluid velocity vector}
\mathbf{U} = \mathbf{\nabla} \phi(x, y, z, t, \theta).
\end{equation}

\subsection{Geomagnetic Formulations}
Based on the hydrodynamic wake expressions obtained from the previous subsection, in what follows we derive the magnetic wake anomaly expressions resulting from the movement of the vessel in finite-depth water. Let $(\sigma_{\mathrm{b}}, \varepsilon_{\mathrm{b}}, \mu_{\mathrm{b}})$, $(\sigma_{\mathrm{w}}, \varepsilon_{\mathrm{w}}, \mu_{\mathrm{w}})$ and $\sigma_{\mathrm{a}}, \varepsilon_{\mathrm{a}}, \mu_{\mathrm{a}})$ be the magnetic permeability coefficient, the dielectric coefficient, and the electrical conductivity coefficient for $z \leq -d$ (i.e., seabed), $-d<z \leq 0$ (i.e., water), and $z>0$ (i.e., air) respectively. The Maxwell equations for electric and magnetic fields $\mathbf{E}$ and $\mathbf{B}$, together with the Ohm law for electrical conductors form the geomagnetic equations relating to the three aforementioned layers are formally expressed as \eqref{eq:Maxwell’s equation} shown at the bottom of this page, 
where $\rho_{\mathrm{e}}$ denotes the electrical flux density, $\mathbf{B}_T = \mathbf{B} + \mathbf{B}_E$, in which $\mathbf{B}_E$ is the geomagnetic field and $\mathbf{B}$ represents the magnetic wake field. In practice, we can approximate $\mathbf{B}_T \approx \mathbf{B}_E$ with high precision due to the fact that the amplitude of the geomagnetic field is much larger than that of the induced magnetic wake. Considering this, together with expressions in \eqref{eq:velocity potential function} and \eqref{eq:fluid velocity vector}, the electric and magnetic harmonic fields corresponding to the solution of Maxwell's equation \eqref{eq:Maxwell’s equation} can be written as
\addtocounter{equation}{1}
\begin{subequations}
\label{eq:opt_power}
    \begin{align}
\mathbf{H} &= \mathbf{h}(\theta, z) e^{-i (\omega_0 t + k_0 x \cos(\theta) + k_0 y \sin(\theta))} 
\\
\mathbf{E} &= \mathbf{e}(\theta, z) e^{-i (\omega_0 t + k_0 x \cos(\theta) + k_0 y \sin(\theta))} 
    \end{align}
\end{subequations}
wherein the magnetic harmonic component $\mathbf{h}(\theta, z)$ is obtained as
\begin{multline}
\label{eq:magnetic harmonic component}
\mathbf{h}(\theta, z) = \mathbf{h}_a e^{-\beta_a z} \tau(z) + \mathbf{h}_b e^{\beta_b (z+d)} \tau(-z-d) \\
+ \left[ \mathbf{h}^{(w^+)}  e^{\beta_w z} + \mathbf{h}^{(w^-)}  e^{-\beta_w z} + \hat{\mathbf{h}}^{(w^+)}  e^{k_0 z} + \hat{\mathbf{h}}^{(w^-)}  e^{-k_0 z} \right] \times
\\
\tau(z+d) \tau(-z)
\end{multline}
in which $\tau(z)$ is the unit step function. The first and the second terms formulate the magnetic harmonic in the air and seabed respectively, and the third term corresponds to the movement of two waves in the fluid in opposite directions. All unknown coefficients in \eqref{eq:magnetic harmonic component} are computed by applying boundary conditions. After some mathematical manipulations, one can verify that the unknown parameters $\mathbf{h}_a$ , $\mathbf{h}_b$ , $\mathbf{h}^{(w^+)} $ , $\mathbf{h}^{(w^-)} $ , $\hat{\mathbf{h}}^{(w^+)} $ , $\hat{\mathbf{h}}^{(w^-)} $ are found as follows

\begin{subequations}
\label{eq:betas}
    \begin{align}
\beta_a^2 &= k_0^2 - \varepsilon_{\mathrm{a}} \mu_{\mathrm{a}} \omega_0^2
\\
\beta_w^2 &= k_0^2 - \varepsilon_{\mathrm{w}} \mu_{\mathrm{w}} \omega_0^2 - i \sigma_{\mathrm{w}} \mu_{\mathrm{w}} \omega_0
\\
\beta_b^2 &= k_0^2 - \varepsilon_{\mathrm{b}} \mu_{\mathrm{b}} \omega_0^2 - i \sigma_{\mathrm{b}} \mu_{\mathrm{b}} \omega_0
    \end{align}
\end{subequations}

\begin{align}
\label{eq:hs}
\hat{\mathbf{h}}^{(w^+)} &= \frac{k_0 \sigma_{\mathrm{w}}}{2 (k_0^2 - \beta_w^2)} \times \frac{e^{k_0 d}}{\cosh(k_0 d)} \times 
\notag
\\
&
\left(\mathbf{B}_E \cdot (i \cos(\theta) \mathbf{a}_x + i \sin(\theta) \mathbf{a}_y - \mathbf{a}_z) \right)  \times
\notag
\\
&
(i \cos(\theta) \mathbf{a}_x + i \sin(\theta) \mathbf{a}_y + \mathbf{a}_z)
\end{align}

\begin{align}
\label{eq:hs}
\hat{\mathbf{h}}^{(w^-)} &= \frac{k_0 \sigma_{\mathrm{w}}}{2 (k_0^2 - \beta_w^2)} \times \frac{e^{k_0 d}}{\cosh(k_0 d)} \times 
\notag
\\
&
\left(\mathbf{B}_E \cdot (i \cos(\theta) \mathbf{a}_x + i \sin(\theta) \mathbf{a}_y + \mathbf{a}_z) \right)  \times
\notag
\\
&
(i \cos(\theta) \mathbf{a}_x + i \sin(\theta) \mathbf{a}_y - \mathbf{a}_z)
\end{align}

The z-components of $\mathbf{h}^{(w^+)} $ and $\mathbf{h}^{(w^-)} $ are obtained as follows:

\begin{multline}
\hspace{-12pt}\left[
\begin{array}{c}
    h_z^{(w^+)} \\
    h_z^{(w^-)}
\end{array}
\right] = 
\left[
\begin{array}{cccc}
    e^{-\beta_w d} \left(\frac{\beta_w}{\beta_b} - \frac{\mu_{\mathrm{w}}}{\mu_{\mathrm{b}}}\right) & -e^{\beta_w d} \left(\frac{\beta_w}{\beta_b} + \frac{\mu_{\mathrm{w}}}{\mu_{\mathrm{b}}}\right) \\
    \left(\frac{\beta_w}{\beta_b} + \frac{\mu_{\mathrm{w}}}{\mu_{\mathrm{a}}}\right) & -\left(\frac{\beta_w}{\beta_b} - \frac{\mu_{\mathrm{w}}}{\mu_{\mathrm{a}}}\right)
\end{array}
\right]^{-1} \\
\times\left[
\begin{array}{c}
    \hat{h}_z^{(w^+)} e^{-k_0 d} \left(\frac{\mu_{\mathrm{w}}}{\mu_{\mathrm{b}}} - \frac{k_0}{\beta_b}\right) + \hat{h}_z^{(w^-)}  e^{k_0 d} \left(\frac{\mu_{\mathrm{w}}}{\mu_{\mathrm{b}}} + \frac{k_0}{\beta_b}\right) \\
    -\hat{h}_z^{(w^+)}  e^{-k_0 d} \left(\frac{\mu_{\mathrm{w}}}{\mu_{\mathrm{a}}} - \frac{k_0}{\beta_a}\right) - \hat{h}_z^{(w^-)}  e^{k_0 d} \left(\frac{\mu_{\mathrm{w}}}{\mu_{\mathrm{a}}} - \frac{k_0}{\beta_a}\right)
\end{array}
\right]
\end{multline}

The air and seabed harmonics are obtained respectively as:

\begin{subequations}
\begin{align}
h_x^a &= i \cos(\theta) \left(\frac{\beta_b}{k_0}\right) h_z^a 
\\
h_y^a &= i \sin(\theta) \left(\frac{\beta_a}{k_0}\right) h_z^a
\\
h_z^a &= -\frac{\beta_w}{\beta_a} \left(h_z^{(w^+)}  - h_z^{(w^-)} \right) - \frac{k_0}{\beta_a} \left(\hat{h}_z^{(w^+)}  - \hat{h}_z^{(w^-)} \right) 
\\
h_x^b &= i \cos(\theta) \left(\frac{\beta_b}{k_0}\right) h_z^b 
\\
h_y^b &= i \sin(\theta) \left(\frac{\beta_b}{k_0}\right) h_z^b 
\\
h_z^b &= -\frac{\beta_w}{\beta_b} \left(e^{-\beta_w d} h_z^{(w^+)}  - e^{\beta_w d} h_z^{(w^-)} \right) 
\notag \\
& \hspace{45pt}
- \left(\frac{k_0}{\beta_b}\right) \left(e^{-k_0 d} \hat{h}_z^{(w^+)}  - e^{k0 d} \hat{h}_z^{(w-) }\right)
\end{align}
\end{subequations}
The $x$ and $y$ components of  $\mathbf{h}^{(w^+)} $ and $\mathbf{h}^{(w^-)} $ are found as 
\begin{multline}
\left[
\begin{array}{cc}
    \begin{array}{c}
        h_x^{(w^+)}  \\
        h_x^{(w^-)} 
    \end{array} \\
    \begin{array}{c}
        h_y^{(w^+)}  \\
        h_y^{(w^-)} 
    \end{array}
\end{array}
\right] = 
\left[
\begin{array}{cccc}
    1 & 1 & 0 & 0 \\
    0 & 0 & 1 & 1 \\
    e^{-\beta_w d} & e^{\beta_w d} & 0 & 0 \\
    0 & 0 & e^{-\beta_w d} & e^{\beta_w d}
\end{array}
\right]^{-1} \\
\times\
\left[
\begin{array}{c}
    h_x^a - \hat{h}_x^{(w^+)}  - \hat{h}_x^{(w^-)}  \\
    h_y^a - \hat{h}_y^{(w^+)}  - \hat{h}_y^{(w^-)}  \\
    h_x^b - e^{-k_0 d} \hat{h}_x^{(w^+)}  - e^{k_0 d} \hat{h}_x^{(w^-)}  \\
    h_y^b - e^{-k_0 d} \hat{h}_y^{(w^+)}  - e^{k_0 d} \hat{h}_y^{(w^-)} 
    \end{array}
    \right]
\end{multline}

Finally, the magnetic wake is obtained as follows:
\begin{align}
\mathbf{H}(x,y,z,t) &= \mathcal{R}e \bigg\{ \int_{-\pi/2}^{\pi/2} \mathbf{h}(\theta, z) A(\theta) \kappa({\theta})  \nonumber\\
& \times e^{-i(\omega_0 t + k_0 x \cos(\theta) + k_0 y \sin(\theta))} \, d\theta
\bigg\}.
\label{eq:long-formula}
\end{align}
\begin{remark}
Although all computed expressions are presented here for the general case of finite-depth water of depth $d$, they can also be applied to deep water by considering $d \rightarrow \infty$. This can be validated by comparing the obtained equations to the corresponding ones represented for deep water expressed in \cite{madurasinghe1994induced}.
\end{remark}

\section{Problem Formulation}
We consider a detection scenario according to Fig. \ref{fig:airborne sensor} wherein a magnetic sensor is positioned at point $(x_0, y_0, h_0)$ at time $t=0$ where $h_0$ is the altitude of the sensor. The magnetic sensor travels at a constant velocity $V_0$ with  angle $\alpha$ relative to the $x$-axis, sampling the magnetic wake at rate $f_0$. The instantaneous location of the sensor at time $t$,  is then formulated as follows:
\begin{subequations}
\label{eq:sensor position}
\begin{align}
x &= x_0 + (V_0 \cos \alpha - V_s) t \\
y &= y_0 + (V_0 \sin \alpha) t \\
z &= h_0
\end{align}
\end{subequations}

Considering a vessel with length $L_s$, beam $S_0$, and draft $h_s$, the vessel hull is commonly modeled as one of the following forms:
\\
Wigley Hull:
\begin{subequations}
\label{eq:vessel hull}
\begin{align}
\label{eq:wigley hull}
S(L_s, x, y) &= S_0 \left[ 1 - \left( \frac{x}{L_s/2} \right)^2 \right] \left[ 1 - \left( \frac{y}{h_s} \right)^2 \right] \nonumber\\
x &\in \left[ -\frac{L_s}{2}, \frac{L_s}{2} \right], \nonumber\\
y &\in \left[ -h_s, 0 \right],
\end{align}
Parabolic hull:
\begin{align}
S(L_s, x, y) &= S_0 \left[ 1 - \left( \frac{x}{L_s/2} \right)^2 \right]\nonumber\\
x &\in \left[ -\frac{L_s}{2}, \frac{L_s}{2} \right],\nonumber\\
y &\in \left[ -h_s, 0 \right],
\end{align}
Prolate Spheroid Hull:
\begin{align}
\label{eq:Prolate hull}
S(L_s, x, y) &= \sqrt{ S_0 \left[ 1 - \left( \frac{x}{L_s/2} \right)^2 \right] - \left( \frac{y}{h_s} \right)^2 }\nonumber\\
x &\in \left[ -\frac{L_s}{2}, \frac{L_s}{2} \right], \nonumber\\
y &\in \left[ -h_s, 0 \right],
\end{align}
\end{subequations}
Let's $\Pv\in\Dv$ denote the vessel parameters, defined as follows:
\begin{align}
    \label{eq:tye6378}
    \Pv 
    &= [ L_s, h_s, S_0, V_s, \alpha ],
\end{align}
where $\Dv$ is the corresponding domain for $\Pv$.
Noting that the velocity of the movement of the magnetic sensor is generally much higher than the velocity of the vessel (i.e., $V_0 \gg V_s$), we can obtain the position-independent form of the Kelvin electromagnetic wave captured by the sensor as follows:
\begin{multline}
\label{eq:Kelvin wave captured}
\mathbf{H}(t;\Pv) = \\
\mathcal{R}e \bigg\{ \int_{-\pi/2}^{\pi/2}\mathbf{h}(\theta, h_0) A(\theta) \kappa({\theta})
e^{-i(\omega_1 t + \omega_2 t + \omega_3)} d\theta\bigg\}
\end{multline}
where
\begin{subequations}
\begin{align}
\omega_1 &= \omega_0 - k_0 V_s \cos(\theta) \\
\omega_2 &= k_0 \cos(\theta - \alpha) \\
\omega_3 &= k_0 x_0 \cos(\theta) + k_0 y_0 \sin(\theta). 
\end{align}
\end{subequations}
By substituting \eqref{eq:sensor position} in \eqref{eq:Kelvin wave captured}, we can write
\begin{multline}
\label{eq:main_eq}
\mathbf{H}(t;\Pv) = \mathcal{R}e \bigg\{ \frac{V_s}{2\pi} \int_{-\pi/2}^{\pi/2} \mathbf{h}(\theta, h_0) A(\theta) 
\times \\
\left[\iint_{S'} \frac{\partial S}{\partial x'} \cosh(k_0 (z' + d)) e^{-i  k_0 (x' \cos(\theta) + y' \sin(\theta))} dx' dy' \right]
\\
\times e^{-i(\omega_1 t + \omega_2 t + \omega_3)} d\theta\bigg\},
\end{multline}
Equation \eqref{eq:main_eq} demonstrates the relationship between the magnetic field measured by the airborne sensor $\boldsymbol{H}(t;\Pv)$ and the geometry of the vessel hull $S$. 
Now $\boldsymbol{H}(t;\Pv)$ can be expressed for $K$ successive samples as follows:
\begin{align}
    \label{eq:main_int_eq_section1}
    \mathbf{H}(t;\Pv)= \mathcal{R}e \left\{ \int_{-\pi/2}^{\pi/2} \mathbf{G}(\theta,t;\Pv) F(\theta;\Pv) d\theta \right\}, 
    \forall t\in\mathcal{T},
\end{align}
where $\mathcal{T}=\left\{0,\frac{1}{f_0},\frac{2}{f_0},..., \frac{K-1}{f_0}\right\}$, and
\begin{subequations}
    \begin{align}
    \label{eq:main_int_eq_section2}
    \mathbf{G}(\theta,t;\Pv) &= \frac{V_s}{2\pi} \mathbf{h}(\theta, h_0) A(\theta) e^{-i(\omega_1 t + \omega_2 t + \omega_3)} \\
    \label{eq:main_shape_eq}
    F(\theta;\Pv) &=
    \notag\\
    & \hspace{-15pt}
    \iint_{S'} \frac{\partial S}{\partial x'} \cosh(k_0 (z' + d)) e^{-i  k_0 (x' \cos(\theta) + y' \sin(\theta))} dx' dy'
    \end{align}
\end{subequations}
Equation \eqref{eq:main_shape_eq} simplifies when applied to the shapes of common vessel hulls. For instance, in the case of the Parabolic hull, after some mathematical manipulations, we can write

\begin{align}
F(\theta;\Pv) = & \left[ 4{V_s} S_0 e^{k_0 d} (1 - e^{-k_0 h_s}) + 4{V_s} S_0 e^{-k_0 d} (1 - e^{k_0 h_s})\right] \nonumber\\ \times 
& \left[ \frac{\cos(\delta/2)}{\delta} - \frac{2\sin(\delta/2)}{\delta^2} \right],
\end{align}
in which $\delta = k_0 L_s \cos(\theta)$.

The determination of the vessel length $Ls\in\Pv$ from its magnetic wake, as presented in  \eqref{eq:main_int_eq_section1}, involves the known quantity $\mathbf{H}(t;\Pv)$ on the left-hand side,, while the integrand incorporates the unknown parameter $Ls\in\Pv$, thereby forming an integral equation. \textit{While the primary focus of this paper is the estimation of vessel length $Ls$ through its magnetic wake analysis, it inherently yields a simultaneous determination of all relevant physical parameters of the vessel, $\Pv$.}

\section{Proposed Solution Scheme}

Estimation of vessel parameters from measured magnetic fields necessitates solving the nonlinear integral equations given by \eqref{eq:main_int_eq_section1}. Conventional numerical methods are often ill-suited to such problems due to their sensitivity to initial conditions, potential for instability, and struggle with the high complexity of large sets of integral equations.  Consequently, a Machine Learning (ML)-based solution scheme is proposed leveraging  deep neural networks (DNNs). DNNs, with their ability to approximate complex functions, handle high-dimensional data, and identify intricate patterns, provide an effective solution for complex integral equations with many unknowns, as is the case in our problem.

\subsection{Deep Residual Neural Network (DRNN)}
Solving integral equations with DNNs is challenging as network depth increases, leading to vanishing gradients, gradient explosion, and model degradation. These issues hinder optimization and limit performance. DRNNs address these challenges using shortcut connections that facilitate gradient flow during backpropagation, mitigating the vanishing gradient problem in regular DNNs and enabling stable training of deeper networks. 
\begin{figure}
 	\centering
		\includegraphics [width=150pt]{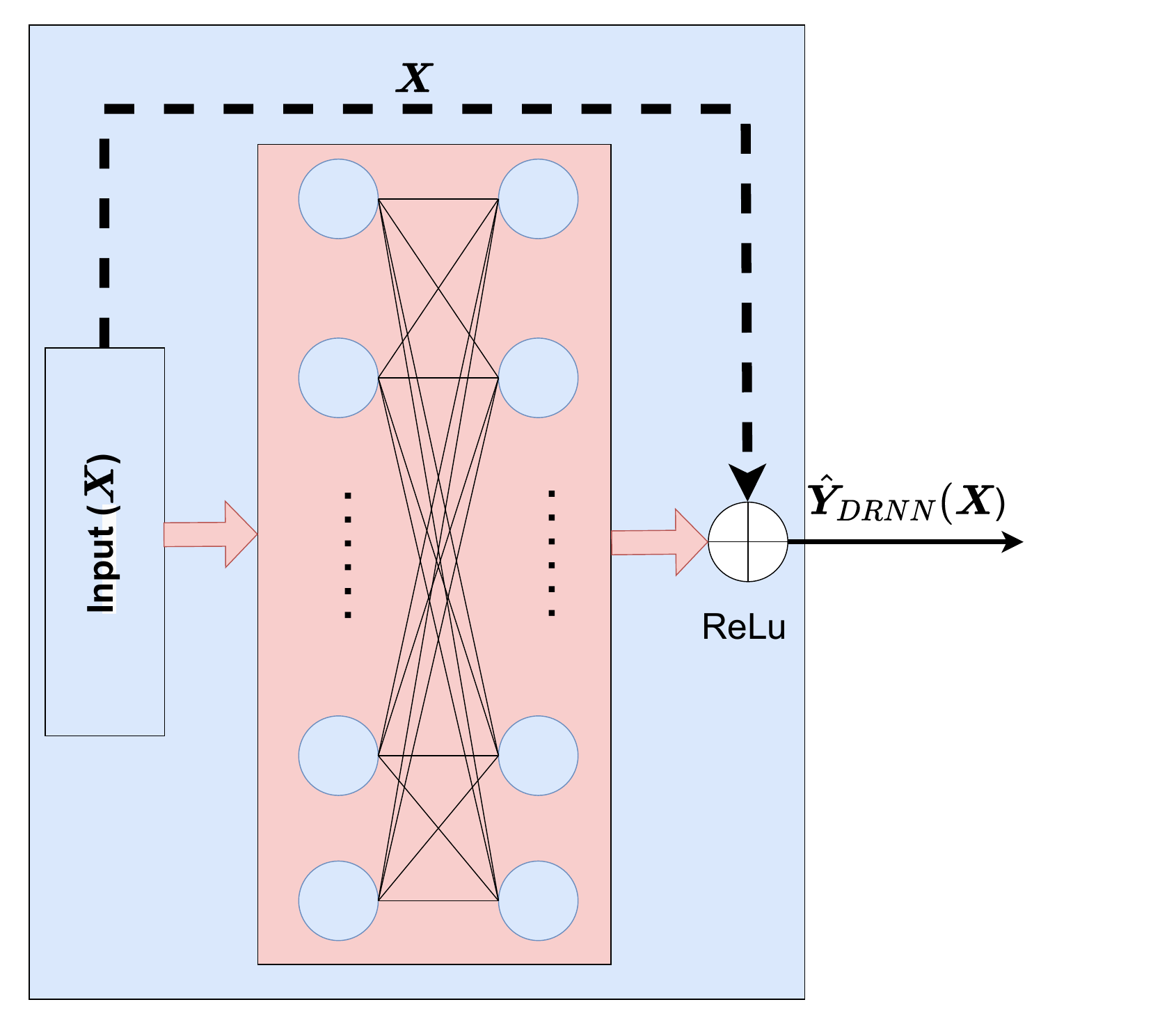} \\
		\caption{Residual block.}
		\label{fig:DRNN block}
\end{figure} 
Fig. \ref{fig:DRNN block} illustrates a single residual block, a fundamental component of DRNNs. This block typically comprises multiple layers, often a combination of layers of neurons, mini-batch normalization, and ReLU activation functions. The core distinguishing feature of DRNNs lies in the skip connection, which adds the input of the residual block, $\boldsymbol{X}$, to the output, $\hat{\boldsymbol{Y}}_{\text{DRNN}}$. 
This enables the network to learn residual mappings, which are the differences between the desired output and the input.   


\subsection{Incorporating DRNN into the Solution Architecture}
\label{sec:traning_DRNN}
In what follows we outline the methodology for employing a typical DRNN consisting of a series of blocks illustrated in Fig. \ref{fig:DRNN block} can be utilized to obtain the physical parameters of the vessel $\Pv$ from its magnetic wake $\boldsymbol{H}(t;\Pv)$. In addition to the vessel parameters defined in \eqref{eq:tye6378}, we consider two categories of vector parameters, including environmental and airborne-sensor vectors denoted by $\Penv\in\Denv$ and $\Psen\in\Dsen$ where $\Denv$ and $\Dsen$ are the domains corresponding to $\Penv$ and $\Psen$, respectively. These vectors are defined as follows:
\begin{subequations}
\begin{align}
    \Penv 
    &= \left[  d, (\varepsilon_{\mathrm{a}}, \mu_{\mathrm{a}}, \sigma_a),(\varepsilon_{\mathrm{w}}, \mu_{\mathrm{w}}, \sigma_{\mathrm{w}}), (\varepsilon_{\mathrm{b}}, \mu_{\mathrm{b}}, \sigma_{\mathrm{b}}) \right]
    \\
    \Psen 
    &= [ V_0, f_0, h_0 ]
\end{align}
\end{subequations}
It is important to highlight that, in contrast to other studies where only specific vessel parameters \(V_s\) and/or \(\alpha\) have been estimated using conventional non-ML-based approaches \cite{monemi2023novel,yang2022study,9420242}, our proposed method introduces a novel approach to obtaining all vessel parameters \(\Pv\) simultaneously within a single unified process by utilizing a DRNN.
To train the DRNN, a labeled training-dataset $\{(\boldsymbol{X},\boldsymbol{Y})\}$ of $\Ns$ samples is constructed. A specific realization of the ground-truth target variable $\Pv$, denoted as $\boldsymbol{Y}$, is selected. The corresponding DRNN input vector $\boldsymbol{X}$  is subsequently constructed leveraging magnetic wake $\boldsymbol{\mathcal{H}}$, conditioned on known parameters $\Penv$, and $\Psen$. Therefore, the input-output labeled data pair corresponding to each sample of the dataset is represented as follows:

%
\begin{align}
\label{eq:DRNN_Input}
(\boldsymbol{X},\boldsymbol{Y})=\left(\underbrace{{\boldsymbol{\mathcal{H}}}, {\Penv}, {\Psen}}_{{\boldsymbol{X}}} ,\underbrace{{\Pv}}_{{\boldsymbol{Y}}}\right)
\end{align}
where
\begin{align}
\label{eq:calH}
  \boldsymbol{\mathcal{H}}
        &=
        \left[\boldsymbol{\|H}(0)\|, \|\boldsymbol{H}({1}/{f_0})\|,...,\|\boldsymbol{H}({(K-1)}/{f_0}\|)\right]\
\end{align}\

In \eqref{eq:DRNN_Input}, we have selected $K$ samples of $\mathbf{H}(t;\Pv)$ scanned by the airborne sensor ($\boldsymbol{\mathcal{H}}$) and concatenated these samples with the known environmental and sensor parameters.
Let $\hat{\boldsymbol{Y}}_{\text{DRNN}}(\boldsymbol{X} , \boldsymbol{\tau})\in\Dv$ be the output of the DRNN corresponding to $\boldsymbol{X} $ and learnable parameter vector $\boldsymbol{\tau}$. 
From \eqref{eq:tye6378}, it is seen that each element of $\hat{\boldsymbol{Y}}_{\text{DRNN}}$ corresponds to a distinct vessel parameter, each possessing its own unique domain, which may differ from the others.  To ensure that each element contributes equitably to the loss function and prevent any single element from disproportionately influencing the overall loss, it is essential to scale the values of each element appropriately. 
Let $f_{\mathrm{Norm}}(\cdot):\Dv\rightarrow [0,1]^{N}$ be a linear vector function mapping the input vector to the corresponding normalized vector,
Our objective is to train the DRNN, such that the following loss function is minimized:
\begin{equation}
    \label{eq:loss_drnn}
    \mathcal{L}_{\text{DRNN}}(\boldsymbol{\tau}) = \mathbb{E}\left\{ \|f_{\mathrm{Norm}}(\boldsymbol{Y}) - f_{\mathrm{Norm}}(\hat{\boldsymbol{Y}}_{\text{DRNN}}(\mathbf{X}, \boldsymbol{\tau}))\|^2 \right\}
\end{equation}
where $\boldsymbol{Y}$ is the ground truth parameter values corresponding to the training data sample vector $\boldsymbol{X}$.
 \begin{figure*}[t]
	\centering
		\includegraphics [width=512pt]{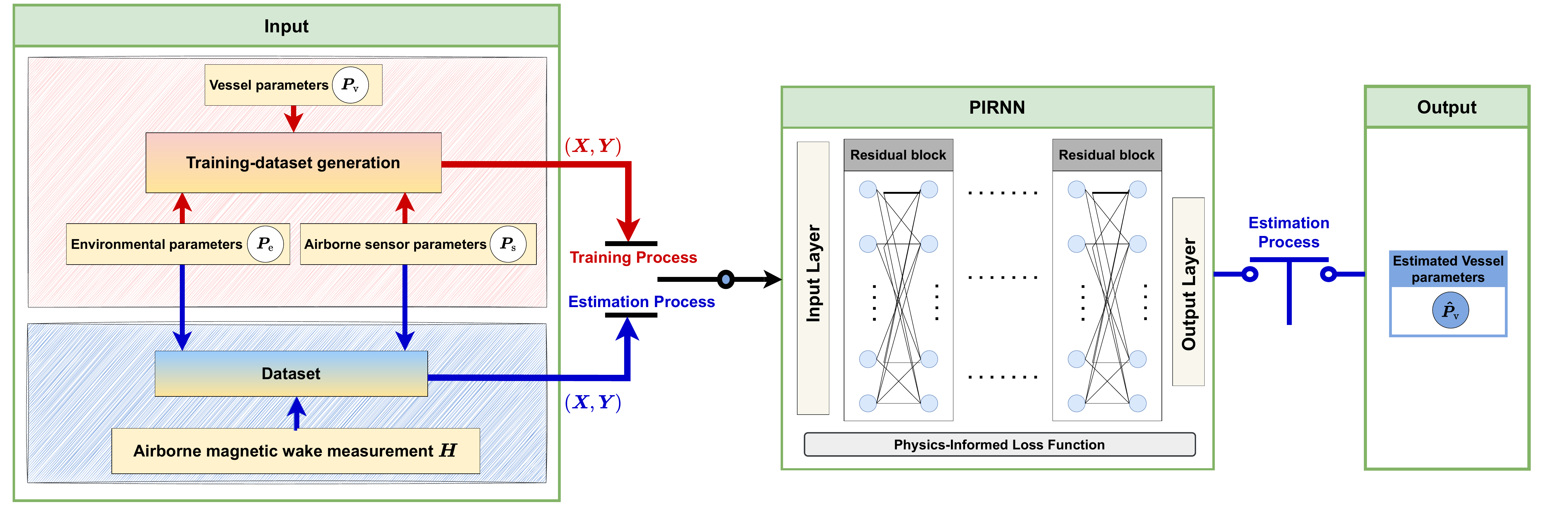} \\
		\caption{ Overall structure of the proposed PIRNN.} 
		\label{fig:resnet}
\end{figure*}

\subsection{Physics-Informed Residual Neural Networks (PIRNN): A Refined Solution}

DRNNs outperform NNs in deep learning tasks by addressing the vanishing gradient issue. However, traditional DRNNs are limited by their exclusive reliance on data-driven learning and their inability to incorporate underlying physical laws directly into the modeling process as discussed in section \ref{sec:Numerical Results}. Consequently, this limitation can result in solutions that may lead to rather significant errors as observed in the numerical results.

Physics-Informed Residual Neural Networks (PIRNNs) address this limitation by integrating the governing physical laws directly into the neural network's loss function. This integration ensures that the solutions not only fit the data but also adhere to fundamental physical laws, thereby enhancing their accuracy and reliability for modeling wave propagation problems.  
We formulate the proposed PIRNN loss function as follows:
\begin{align}
    \label{eq:loss_pirnn}
    \mathcal{L}_{\text{PIRNN}}(\boldsymbol{\tau}) 
    &=
    \mathbb{E} \left\{\|{\mathcal{\boldsymbol{\mathcal{H}}}} -         \mathcal{R}e \int_{-\frac{\pi}{2}}^{\frac{\pi}{2}} \boldsymbol{\mathcal{G}}(\theta,\boldsymbol{\tau}) \mathcal{F}(\theta,\boldsymbol{\tau}) d\theta\|^2 \right\}
\end{align}
where $\boldsymbol{\mathcal{H}}$ is given by \eqref{eq:calH}, and $\boldsymbol{\mathcal{G}}(\theta,\boldsymbol{\tau})$, $\mathcal{F}(\theta,\boldsymbol{\tau})$  are expressed as 
\begin{align}
\label{eq:loss_pirnn_param}
        \boldsymbol{\mathcal{G}}(\theta,\boldsymbol{\tau})
        &=
        \left[\boldsymbol{G}(\theta, 0;\hat{\Pv}(\boldsymbol{\tau})),... ,\boldsymbol{G}(\theta, \frac{K-1}{f_0};\hat{\Pv}(\boldsymbol{\tau}))\right]\bigg|_{\Penv,\Psen}
        \notag
        \\
    \mathcal{F}(\theta,\boldsymbol{\tau})
        &=
        \left[F(\theta,0;\hat{\Pv}(\boldsymbol{\tau})),...,F(\theta,\frac{K-1}{f_0};\hat{\Pv}(\boldsymbol{\tau}))\right]\bigg|_{\Penv,\Psen}
\end{align}
In \eqref{eq:loss_pirnn_param},  $\hat{\Pv}(\boldsymbol{\tau})$ denotes the estimation of $\Pv$ which is obtained as the output of the PIRNN using the ground truth sample data $\boldsymbol{\mathcal{H}}$ as input. The corresponding physics-informed estimation of $\boldsymbol{\mathcal{H}}$ is denoted by the term   $\mathcal{R}e \int_{-\frac{\pi}{2}}^{\frac{\pi}{2}} \boldsymbol{\mathcal{G}}(\theta,\boldsymbol{\tau}) \mathcal{F}(\theta,\boldsymbol{\tau}) d\theta$  in \eqref{eq:loss_pirnn}. This term depends on the output of PRINN, $\hat{\Pv}(\boldsymbol{\tau})$ as seen in \eqref{eq:loss_pirnn_param}. 
We adopt the Monte Carlo method to evaluate the integral term in \eqref{eq:loss_pirnn}. This approach is tailored to strengths and adaptability in calculating complex, high-dimensional integral. 
Its reliance on random sampling enables accurate estimations even when dealing with intricate, irregular, or discontinuous functions, obviating the need for complex mathematical transformations. 
The integral term in \eqref{eq:loss_pirnn} can be calculated based on the Monte Carlo method as follows:
\begin{multline}
\label{eq:mont-carlo}
\int_{-\frac{\pi}{2}}^{\frac{\pi}{2}} \boldsymbol{\mathcal{G}}(\theta,\boldsymbol{\tau}) \mathcal{F}(\theta,\boldsymbol{\tau}) d\theta
\approx 
\\
\frac{\pi}{N_{\text{mont-carlo}}} \sum_{j=1}^{N_{\text{mont-carlo}}} \boldsymbol{\mathcal{G}}(\theta_j,\boldsymbol{\tau}) \mathcal{F}(\theta_j,\boldsymbol{\tau}), \quad \boldsymbol{\theta}_j \sim \mathrm{U}\left(-\frac{\pi}{2}, \frac{\pi}{2}\right)
\end{multline}
where $\mathrm{U}(\cdot,\cdot)$ is the uniform probability distribution function and $N_{\text{mont-carlo}}$ denotes the number of samples randomly selected within the interval of integration.



Two observations are noteworthy in this context. Firstly, it is seen that the values of the environmental and sensor parameters, $\Penv$ and $\Psen$, are directly incorporated into the computation of the PIRNN loss function expressed in \eqref{eq:loss_pirnn}. In contrast, these parameters are not utilized in the loss function of DRNN in \eqref{eq:loss_drnn}.
Secondly, the non-physics-informed loss function in \eqref{eq:loss_drnn} directly depends on the vessel length $\Ls$ as one of the elements of $\Pv$, whereas  the physics-informed loss function in \eqref{eq:loss_pirnn}  depends on the magnetic field $\boldsymbol{\mathcal{H}}$. To compare the efficiency of these two loss functions within the learning algorithm, we propose the error function for the learning algorithm, which is formulated as follows:
\begin{align}
\label{eq:final error}
    \mathcal{E(\boldsymbol{\tau})}= \frac{1}{\Nb} \sum_{i=1}^{\Nb} \|L_i - \hat{L}_i( \boldsymbol{\tau})\|^2
\end{align}
where $\Nb<\Ns$ is the size of  training mini-batch that sampled from training-dataset of size $\Ns$ and $L_i$ is the true vessel length $\Ls$ of the $i$'th sample in the mini-batch dataset and $\hat{L}_i( \boldsymbol{\tau})$ is the corresponding vessel length estimated using DRNN/PIRNN.

\begin{algorithm}
\caption{DRNN/PIRNN training process}
\begin{algorithmic}[1]
\Statex \hspace{-20pt} \textbf{Input Parameters:} 
        \State $\{(\boldsymbol{X},\boldsymbol{Y})\}$: Training-dataset
        \State $M$: Number of training iterations

\Statex \hspace{-20pt} \textbf{Output Parameters:} 
    \State $\boldsymbol{\tau}$: Learnable Parameters of the residual neural network for the estimation of  the vessel parameters $\hat{\Pv}(\boldsymbol{\tau})$.
\Statex \hspace{-20pt} \textbf{Initialization:} 
\State Initialize the neural network parameter $\boldsymbol{\tau}$ randomly.
  \Statex \hspace{-20pt} \textbf{Training:}
\For{$m = 1$ to $M$}
    \State \textbf{Mini-batch Sampling:} Select a random mini-batch $\boldsymbol{Z}_{\mathrm{mini-batch}}$ containing $\Nb$ training samples $\boldsymbol{Z}=(\boldsymbol{X},\boldsymbol{Y})$ from the training-dataset.
    
    \State \textbf{Forward Propagation:} 
    \State \hspace{10pt} {\it DRNN}:
    For each data sample $\boldsymbol{Z}\in\boldsymbol{Z}_{\mathrm{mini-batch}}$, propagate the training data through the neural network to obtain the output $\hat{\boldsymbol{Y}}_{\text{DRNN}}(\boldsymbol{X}, \boldsymbol{\tau})$  in \eqref{eq:loss_drnn}.
    \State \hspace{10pt} {\it PIRNN}:
     For each data sample $\boldsymbol{Z}\in\boldsymbol{Z}_{\mathrm{mini-batch}}$, calculate $\mathcal{R}e \int_{-\frac{\pi}{2}}^{\frac{\pi}{2}} \boldsymbol{\mathcal{G}}(\theta,\boldsymbol{\tau}) \mathcal{F}(\theta,\boldsymbol{\tau}) d\theta$  in \eqref{eq:loss_pirnn}  corresponding to $\boldsymbol{Z}$, using Monte-Carlo approximation through \eqref{eq:mont-carlo}. 
    
    \State \textbf{Loss Computation:} Minimize the loss function \eqref{eq:loss_drnn} (for the DRNN) or \eqref{eq:loss_pirnn} (for the PIRNN) using an optimization algorithm like Adam.
 \State  \textbf{Backward Propagation:} Update the learnable parameter $\boldsymbol{\tau}$ based on the calculated gradients.
\EndFor
\end{algorithmic}
\end{algorithm}

\subsection{Training-Dataset Generation}
\label{sec:training-dataset}
To effectively train the PIRNN, it is essential to obtain a test set for the vessel magnetic wake under various values of $\Penv$, $\Psen$ and $\Pv$. Let $\Denvs \subset \Denv$, $\Dsens \subset \Dsen$, and $\Dvs \subset \Dv$ be the discretized subdomains leveraged to generate training-dataset. To create a complete training-dataset, we need to calculate the magnetic wake $\tilde{\boldsymbol{\mathcal{H}}}$ corresponding to  $\tilde{\boldsymbol{P}}=(\Penvs,\Psens,\Pvs)\in \Denvs \times \Dsens \times \Dvs$. Here, $\tilde{\boldsymbol{\mathcal{H}}}$ is a vector comprising $K$ samples of magnetic wake represented by \eqref{eq:calH} corresponding to $\tilde{\boldsymbol{P}}$, which can be obtained by leveraging \eqref{eq:main_int_eq_section1}, \eqref{eq:main_int_eq_section2} and  \eqref{eq:main_shape_eq}. 
The training-dataset is then formed by aggregating all possible pairs into set  $(\boldsymbol{X},\boldsymbol{Y})=\\(\tilde{\boldsymbol{\mathcal{H}}} ,\tilde{\Pv})|_{\Penv, \Psen}$ as depicted in the red-colored section of the input block in  Fig. \ref{fig:resnet}.

\subsection{Training Algorithm for the Proposed DRNN/PIRNN Structure }
\label{sec:traning_alg}
The training process employed for the estimation of vessel parameters $\hat{\Pv}$ including length $\hat{\Ls}$ is illustrated in Algorithm 1 for both physics-informed and non physics-informed scenarios. The algorithm starts by constituting the training-dataset in Step 2 according to the procedure described in Section \ref{sec:traning_DRNN} and \ref{sec:training-dataset}. We consider the training process through $M$ iterations where in each iteration the neural network is trained using a random mini-batch containing $\Nb$ training samples. At each iteration, after selecting the mini-batch, we need to calculate the right term of the loss function as the forward process corresponding to 
 the terms $\hat{\boldsymbol{Y}}_{\text{DRNN}}(\boldsymbol{X}, \boldsymbol{\tau})$  and $\mathcal{R}e \int_{-\frac{\pi}{2}}^{\frac{\pi}{2}} \boldsymbol{\mathcal{G}}(\theta,\boldsymbol{\tau}) \mathcal{F}(\theta,\boldsymbol{\tau}) d\theta$ in \eqref{eq:loss_drnn} and \eqref{eq:loss_pirnn} for DRNN and PIRNN scenarios receptively. Note that for the PIRNN case, first we need to obtain $\hat{\Pv}$ as the output of the neural network then we incorporate this result to calculate $\int_{-\frac{\pi}{2}}^{\frac{\pi}{2}} \boldsymbol{\mathcal{G}}(\theta,\boldsymbol{\tau}) \mathcal{F}(\theta,\boldsymbol{\tau}) d\theta$ using the Monte-Carlo integral approximation as formulated in \eqref{eq:mont-carlo}. The corresponding loss function is then minimized through the Adam optimizer using backward propagation.


Once the proposed DRNN/PIRNN estimation framework is trained, the evaluation phase can become active. 
This network is designed to deduce vessel parameters $\hat{\Pv}$, with a primary focus on vessel length, $\Ls$, by processing the scanned vessel magnetic wake samples $\boldsymbol{\mathcal{H}}$. The operational sequence of converting the measured samples $\boldsymbol{\mathcal{H}}$ to the data pair $(\boldsymbol{X},\boldsymbol{Y})$ is depicted in the blue-shaded region of Fig. \ref{fig:resnet}.


\begin{figure}
	\centering
		\includegraphics [width=200pt]{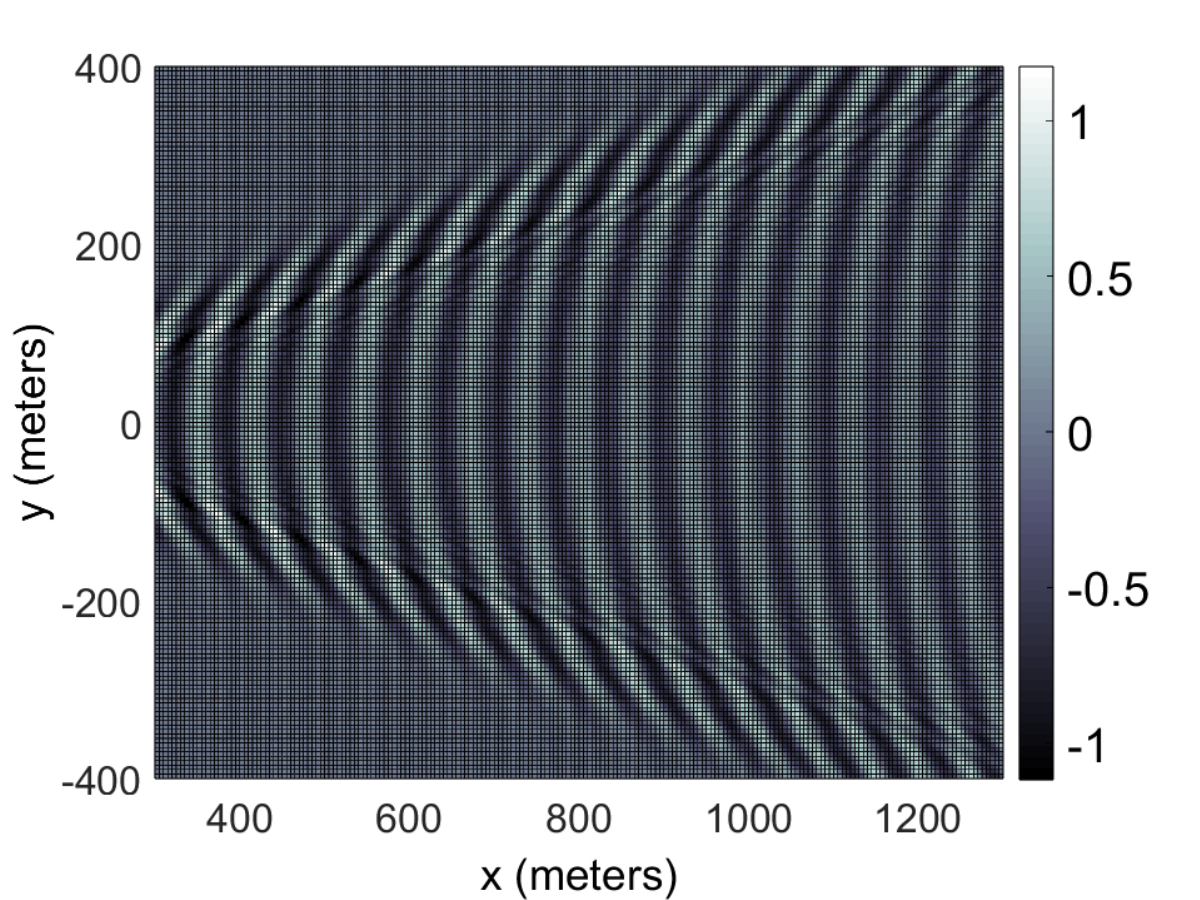} \\
		\caption{ Illustration of the typical vessel magnetic wake  in an area of $800\times1000m^2$.} 
		\label{fig:2D magnetic Wake}
\end{figure}

\section{Numerical Results}
\label{sec:Numerical Results}
In this section, we initially present the parameter values utilized for the training and evaluation of the proposed PRINN-based structure and describe the scenario for generating the vessel magnetic wake across various scenarios. Subsequently, we provide the numerical results of applying and evaluating the proposed PIRNN-based algorithm to estimate vessel parameters based on its magnetic wake. 

\subsection{Parameters Setting and PIRNN Configuration}

As previously mentioned and illustrated in Fig. \ref{fig:airborne sensor}, in each training-dataset generation scenario, the airborne magnetic sensor intersects the vessel's trajectory at an angle $\alpha$ and samples the vessel magnetic wake along this trajectory.
The electromagnetic parameters of $\Penv$ are considered as follows \cite{robert1988electrical}:
\begin{align*}
(\varepsilon_{\mathrm{a}}, \varepsilon_{\mathrm{w}}, \varepsilon_{\mathrm{b}}) &= (\varepsilon_0, 81 \varepsilon_0, 10 \varepsilon_0) \, \text{F/m} \\
(\mu_{\mathrm{a}}, \mu_{\mathrm{w}}, \mu_{\mathrm{b}}) &= (\mu_0, \mu_0, \mu_0) \, \text{NA}^{-2} \\
(\sigma_{\mathrm{a}}, \sigma_{\mathrm{w}}, \sigma_{\mathrm{b}}) &= (0, 5, 0.025) \, \text{S/m}
\end{align*}
Fig. \ref{fig:2D magnetic Wake} shows a typical two-dimensional image of the normalized magnetic wake generated by the movement of a ship, and the Kelvin wave pattern is also clearly visible in this figure. 

To constitute the training-dataset, we have considered a great number of quantized parameter vectors  $\tilde{\boldsymbol{P}}\in \Denvs \times \Dsens \times \Dvs$ where the variation interval of each element of $\tilde{\boldsymbol{P}}$ and the corresponding number of samples per interval is shown in Table \ref{tab:parameters}. 
To form the training-dataset, for each parameter vector $\tilde{\boldsymbol{P}}$, we obtain the corresponding magnetic $\tilde{\boldsymbol{H}}$ as described in Section \ref{sec:training-dataset}.
We consider the Wigley vessel hull model represented in \eqref{eq:wigley hull}. 
The coefficient $\gamma_{\text{Hull}}$ in Table \ref{tab:parameters} has been considered as a fixed parameter, therefore the vessel draft ($\hs$) is not an independent parameter and 
can be excluded from the PIRNN calculations.
The noise is assumed to follow a Gaussian distribution, and the signal-to-noise ratio (SNR) is considered $-10$ dB. 

At each scan of the airborne sensor, we take $K=15$ magnetic wake samples to form the input vector $\boldsymbol{X}$ in \eqref{eq:DRNN_Input}. From  Table \ref{tab:parameters} and \eqref{eq:DRNN_Input}, it is inferred that we need $K+1$ neurons in the input layer of the DRNN/PIRNN, wherein $K$ neurons correspond to the samples of vessel magnetic wake $\boldsymbol{H}$ and one neuron corresponds to the sea depth parameter $d$. 
Subsequent to the input layer, the architecture incorporates an initial sequence of two fully connected layers, each consisting of 64 neurons. This is followed by a cascade of 5 residual blocks. Each residual block is structurally composed of 3 fully connected layers: the first layer with 128 neurons, the second with 256 neurons, and the third with 128 neurons. The Rectified Linear Unit (ReLU) activation function is consistently applied to all layers within these residual blocks. Furthermore, each residual block integrates skip connections that directly link the input to the output, effectively bypassing the intermediate layers. Following the final residual block, the network incorporates a sequence of two fully connected layers, each with 64 neurons. These layers are then succeeded by a final fully connected output layer comprising 4 neurons corresponding to the output parameters $L_s$, $S_0$, $V_s$, and $\alpha$.

The ReLU activation function is employed in all layers. The network is trained using Adam optimizer with a learning rate  $5 \times 10^{-4}$ to minimize the loss function over $M=3000$ iterations where mini-batch size is  $N_b=256$.

\begin{table}
    \centering
    \begin{tabular}{|c|c|c|} \hline 
         \textbf{Parameter}&  \textbf{Parameter interval}& \textbf{Sample per interval}\\ \hline 
         $S_0$&  $[500-5000] \text{ m}^2$& 5\\ \hline 
         $V_s$&  $[1-10] \text{ m}/s$& 10\\ \hline 
         $|\alpha|$&  $[0^\circ-19^\circ]$& 10\\ \hline 
         $L_s$&  $[30-330] \text{ m}$& 150\\ \hline
 $h_s = \gamma_{\text{Hull}} L_s$& $\gamma_{\text{Hull}}\times[30-330] \text{ m}$&-\\\hline
 $V_0$& $20 \text{ m}/s$&1\\\hline
 $f_0$& $10 \text{ Hz}$&1\\\hline
 $h_0$& $50 \text{ m}$&1\\\hline
 $d$& $[100-3000] \text{ m}$&9\\\hline
    \end{tabular}
  \caption{Parameter variation ranges and corresponding sample counts per interval.}
  \label{tab:parameters}
  \end{table}

\begin{figure}
	\centering
		\includegraphics [width=250pt]{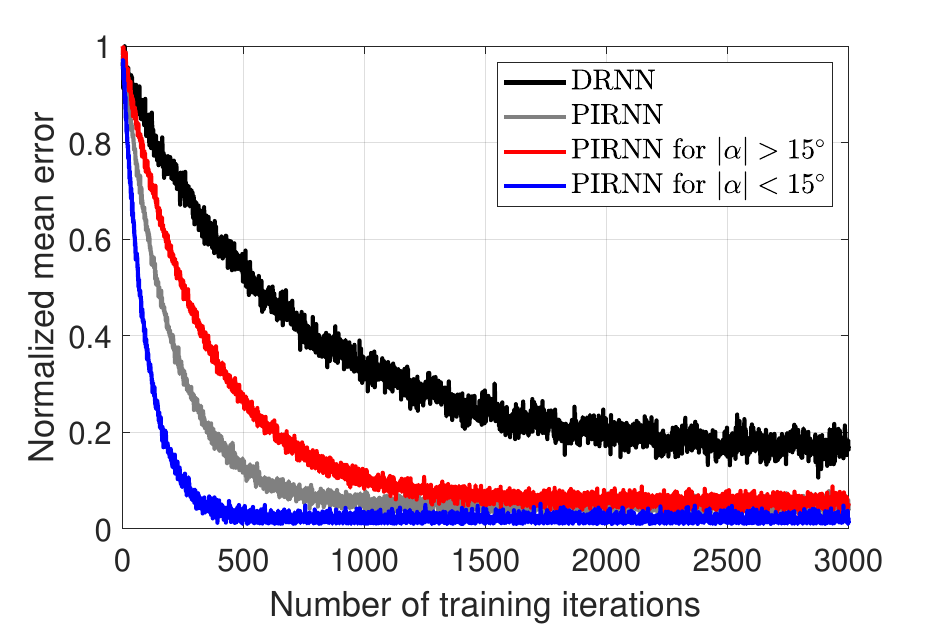} \\
		\caption{Mean error resulting from different training scenario. } 
		\label{fig:resnet_error}
\end{figure}

\begin{figure}
	\centering
		\includegraphics [width=250pt]{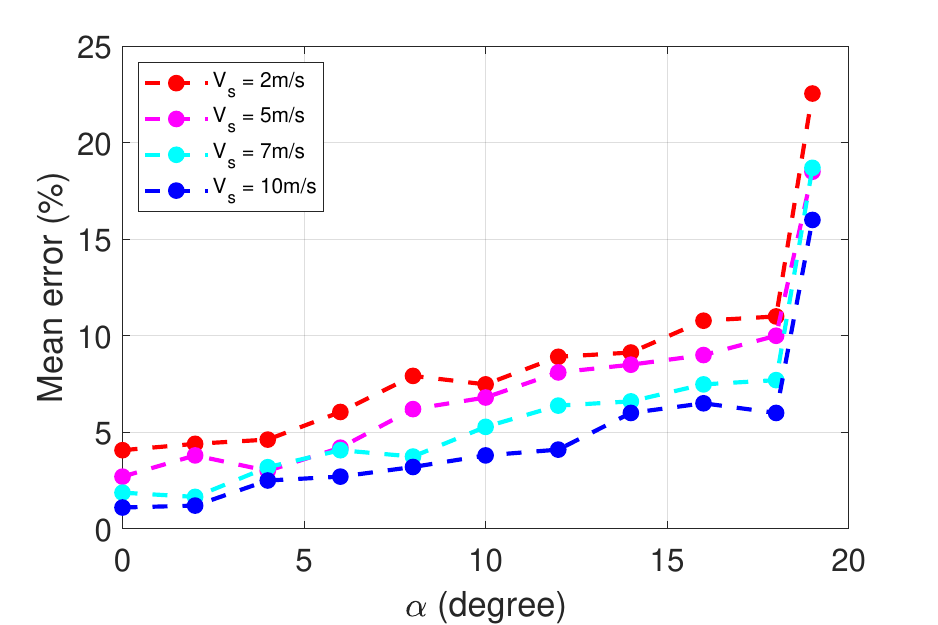} \\
		\caption{Vessel length estimation error vs. airborne scanning angle.} 
		\label{fig:angle_error}
\end{figure} 
\begin{figure}
	\centering
		\includegraphics [width=250pt]{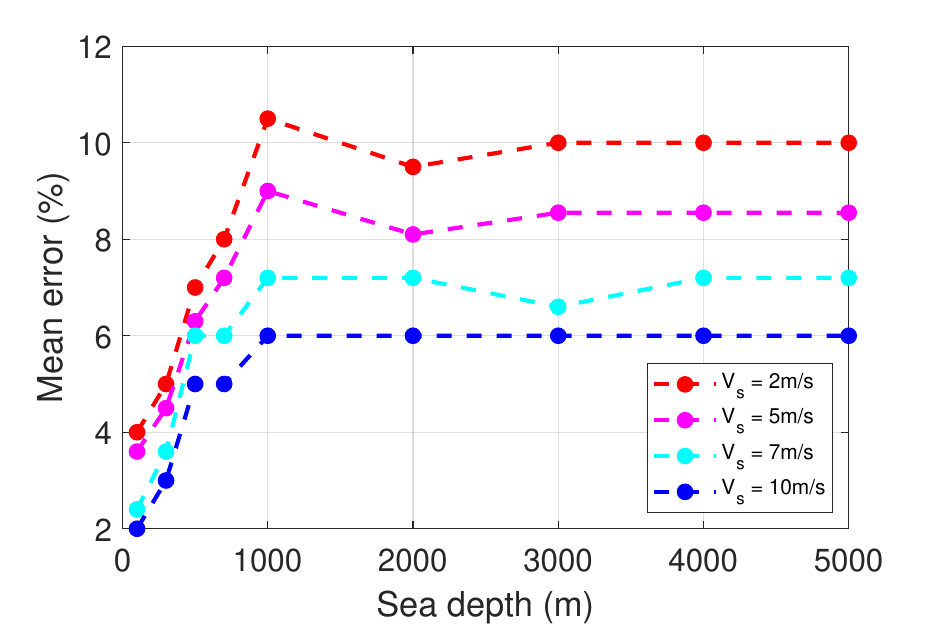} \\
		\caption{Vessel length estimation error vs. sea depth.} 
		\label{fig:depth_error}
\end{figure}

\subsection{Training}

This section details the training process of the proposed machine learning (ML)-based solution using the training-dataset. Figure \ref{fig:resnet_error} presents the normalized error (derived from the non-normalized error in \eqref{eq:final error}) versus training iteration.
For each iteration, a mini-batch of $N_b$ randomly selected data samples is used, encompassing various environmental and vessel parameters as specified in Table \ref{tab:parameters}.
The black curve in Fig. \ref{fig:resnet_error} depicts the normalized error resulting from training with the DRNN loss function, as defined in \eqref{eq:loss_drnn}.  This DRNN implementation exhibits suboptimal performance, characterized by both a slow convergence rate and a error of approximately 20\%.
To address these limitations, we incorporated the governing physical law into the training process within the proposed PIRNN framework.  The resulting performance is illustrated by the non-black curves in Fig. \ref{fig:resnet_error}. Specifically, the gray curve represents the performance obtained by modifying the loss function from \eqref{eq:loss_drnn} to \eqref{eq:loss_pirnn} without altering the original training-dataset.  A substantial increase in convergence rate and a significant reduction in error are observed.
To evaluate the influence of the airborne sensor direction $\alpha$ on the estimation error, we generated modified training-datasets.  These datasets were constructed by selecting mini-batch samples based on the sensor direction $\alpha$ falling within specific intervals: $|\alpha| < 15^\circ$ (corresponding to the blue curve) and $|\alpha| > 15^\circ$ (corresponding to the red curve).
When training samples are selected with a higher incorporation of lower values of $\alpha$, an improvement in performance is observed in terms of training speed and potentially the residual error. Additionally, the convergence of the DRNN, along with the red-colored, gray-colored, and blue-colored PIRNNs, occurs after 2500, 1800, 1000, and 500 iterations, respectively.

\subsection{Evaluation}
Noting that the performance of the DRNN is much lower than that of the PIRNN, in this part we focus more on the evaluation of the proposed PIRNN. 
Fig. \ref{fig:angle_error} illustrates the mean error of the PIRNN of the vessel length estimation $\hat{\Ls}$ obtained from 100 different test input data versus airborne sensor direction $\alpha$ and vessel speed $\Vs$.  It is seen that the mean error in vessel length estimation decreases with increasing vessel speed. This reduction is attributed to the increase in kinetic energy of vessel at higher speeds, which results in a larger amplitude of the magnetic signal received by the airborne sensor. On the hand, it is observed that higher values of $\alpha$ results in greater error. Specifically, as $\alpha$ approaches Kelvin's limit angle $\alpha_{\mathrm{kelvin}}=19.5^\circ$ \cite{Zou2000}, the mean estimation error increases abruptly. This increase is attributed to the insufficient amplitude of the magnetic signal received by the airborne sensor when approaching the critical value $\alpha_{\mathrm{kelvin}}$ as elaborated in \cite{monemi2023novel}.
For a more detailed numerical examination, it is seen from  Fig. \ref{fig:angle_error}  that in the worst-case scenario, at speeds less than 5m/s, the estimation error reaches a maximum of 5\% for $\alpha=0^\circ$ and a maximum of 10\% for $\alpha=15^\circ$, which is considered an acceptable error from a practical perspective. The estimation error is seen to be consistently lower for speeds higher than 5m/s. It should be noted that in practice we may conduct a maneuver for the airborne scanning sensor in a way that small values of $\alpha$ with high values of $\boldsymbol{\mathcal{H}}$ are realized, corresponding to a minimized estimation error. 
Fig. \ref{fig:depth_error} investigates the impact of sea depth $d$ on the vessel length estimation error. It is seen that for $d>1000$m the mean error is persistently constant for a given vessel speed. Conversely, a decrease in sea depth is associated with enhanced estimation accuracy.
 This improvement is attributable to the increased Froude numbers associated with shallower depths \cite{Newman1977}. This phenomenon arises because, at shallower depths, the vessel speed relative to the water depth results in increased resistance and consequently, larger wave amplitudes.

\begin{figure}
	\centering
		\includegraphics [width=250pt]{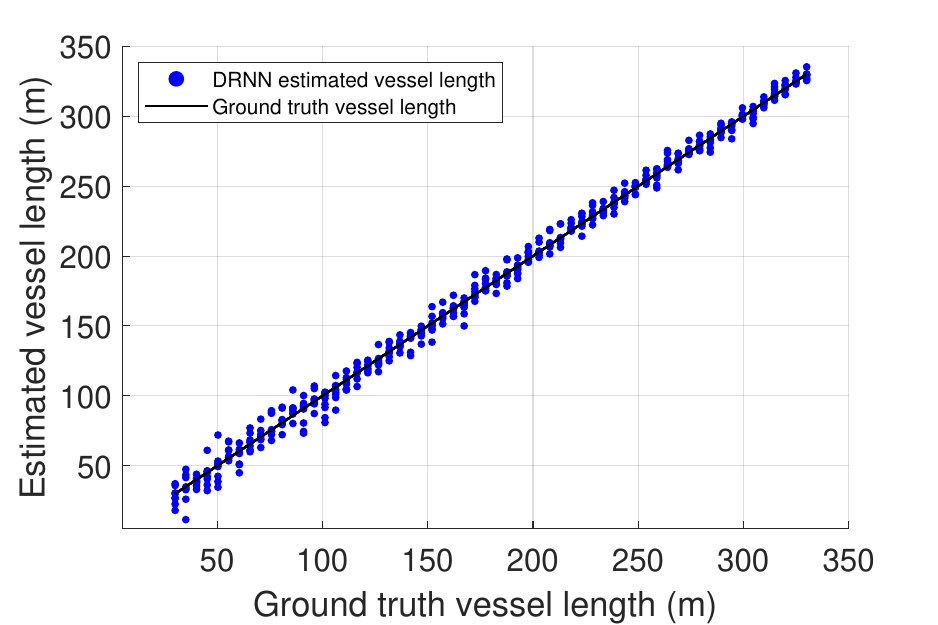} \\
		\caption{Estimated vessel length using DRNN vs. ground truth value.} 
		\label{fig:drnn_length_error}
\end{figure}

\begin{figure}
	\centering
		\includegraphics [width=250pt]{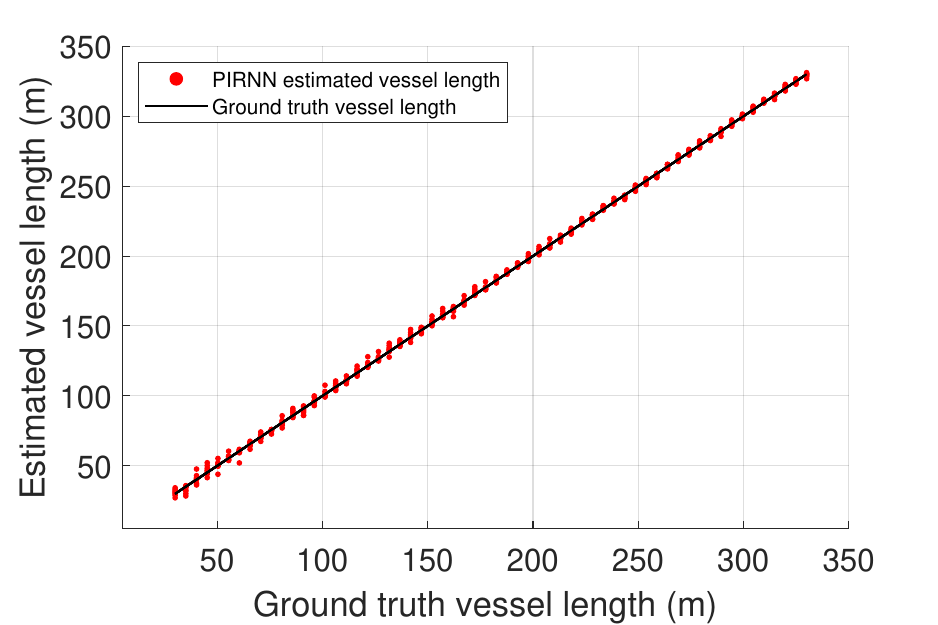} \\
		\caption{Estimated vessel length using PIRNN vs. ground truth value.} 
		\label{fig:pirnn_length_error}
\end{figure} 

Figs. \ref{fig:drnn_length_error} and \ref{fig:pirnn_length_error} present an evaluation of the performance of the proposed DRNN and PIRNN algorithms for estimating vessel length using  its magnetic wake signature. These figures demonstrate the estimated vessel length values versus the corresponding ground truth values $\Ls$, for $\Ls$ ranging from $30$m to $330$m. For each value of $\Ls$, 10 distinct scenarios are chosen whose parameters are selected by considering a random $|\alpha |<15^\circ$ and all parameters other than $\Ls$ and $\alpha$ are randomly selected from the feasible range specified in Table \ref{tab:parameters}.  
It is seen that the PIRNN's output consistently exhibits a small deviation from the corresponding ground truth values across all values of $\Ls$, however, the DRNN suffers from a rather high error. Notably, the error for both methods tends to decrease as  $\Ls$ increases. This improvement is attributed to the increase in kinetic energy of longer vessels, which results in a larger amplitude of the magnetic signal measured by the airborne sensor. For example, it is observed from Fig. \ref{fig:drnn_length_error} and Fig. \ref{fig:pirnn_length_error}  that for the vessel length higher than 100m, the maximum estimation error of
the DRNN and PIRNN solution schemes are about 20\% and 5\%, respectively.


\section{Conclusion}
This research demonstrates the efficacy of fusing physics-based modeling with deep learning for vessel length estimation by leveraging the vessel magnetic wake signatures.  
By deriving nonlinear integral equations that describe the relationship between the magnetic wake and vessel length within a realistic, finite-depth marine environment, and subsequently embedding these hydromagnetic formulations into the loss function of a physics-informed residual neural network (PIRNN), a significant improvement in estimation accuracy was realized compared to using a standard deep residual neural network (DRNN). 
For example, for the vessel length higher than 100m, we have shown that the maximum estimation error among various simulation scenarios is 20\% and 5\% for the proposed DRNN and PIRNN solutions respectively.    
The influence of sea depth on vessel length estimation error was also explored. For depths exceeding 1000m, the mean estimation error remains consistently constant for a given vessel speed. Conversely, a reduction in sea depth results in an improvement in the estimation accuracy.

	\bibliographystyle{IEEEtran}
	\bibliography{Mybib}
	\begin{biography}[{\includegraphics[width=1in,height
=1.25in,clip,keepaspectratio]{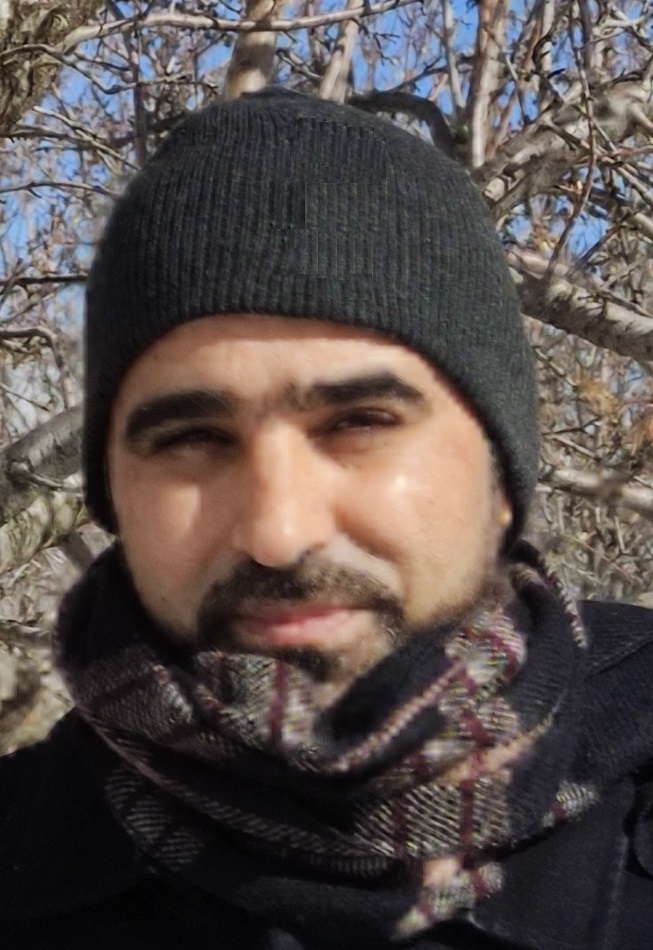}}]{Mohammad Amir Fallah}
		 received the BSc, MSc, and Ph.D. degrees from Shiraz University, Shiraz, Iran, and Tarbiat Modares University, Tehran, Iran, and Shiraz University, Shiraz, Iran, in 2001, 2003 and 2013 respectively, all in electrical and computer engineering. 
    He is an assistant professor with the Department of Engineering, Payame Noor University (PNU), Tehran, Iran, from 2015 till now. His current research interests include antenna and propagation, mobile computing, and the application of machine learning and artificial intelligence in wireless networks.
	\end{biography}
 \vspace{-10pt}
\begin{biography}[{\includegraphics[width=1in,height
=1.25in,clip,keepaspectratio]{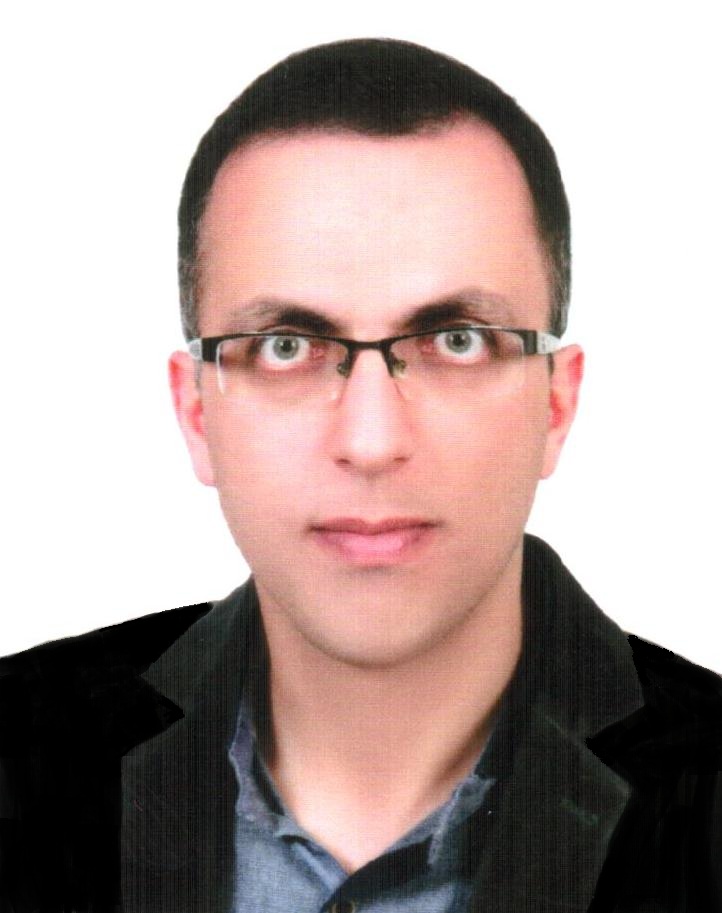}}]{Mehdi Monemi} (Member, IEEE)
received the B.Sc., M.Sc., and Ph.D. degrees all in electrical and computer engineering from Shiraz University, Shiraz, Iran, and Tarbiat Modares University, Tehran, Iran, and Shiraz University, Shiraz, Iran in 2001, 2003 and 2014 respectively. After receiving his Ph.D., he worked as a project manager in several companies and was an assistant professor in the Department of Electrical Engineering, Salman Farsi University of Kazerun, Kazerun, Iran, from 2017 to May 2023. He was a visiting researcher in the Department of Electrical and Computer Engineering, York University, Toronto, Canada from June 2019 to September 2019. He is currently a Postdoc researcher with the Centre
for Wireless Communications (CWC), University of Oulu, Finland. His current research interests include resource allocation in 5G/6G networks, as well as the employment of machine learning algorithms in wireless networks.
	\end{biography}
 \vspace{-10pt}

\begin{biography}[{\includegraphics[width=1in,height=1.25in,clip,keepaspectratio]{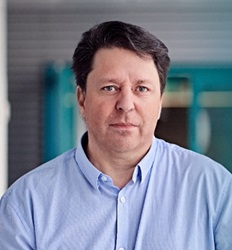}}]{Matti Latva-aho} (Fellow, IEEE) received Dr.Tech. (Hons.) degree in Electrical Engineering from the University of Oulu, Finland, in 1998. Prof. Latva-aho served as the Director of Centre for Wireless Communications (CWC) from 1998 to 2006 and later as Head of the Department of Communication Engineering until August 2014. Currently, he is the Director of the National 6G Flagship Programme and Global Fellow at The University of Tokyo. He has published over 600 conference and journal publications.
\end{biography}

\end{document}